\documentclass[conference]{IEEEtran}
\IEEEoverridecommandlockouts
\usepackage{graphicx} 
\usepackage{amsmath}
\usepackage{xcolor}
\usepackage{multirow}

\usepackage{xcolor}
\usepackage{multirow}
\usepackage{tabularx}
\usepackage{adjustbox}
\usepackage{comment}

\usepackage{algorithm}
\usepackage{algpseudocode}
\algnewcommand\algorithmicforeach{\textbf{for each}}
\algdef{S}[FOR]{ForEach}[1]{\algorithmicforeach\ #1\ \algorithmicdo}

\title{Custom Algorithm-based Fault Tolerance for Attention Layers in Transformers}
\author{
\IEEEauthorblockN{ Vasileios Titopoulos, Kosmas Alexandridis, Giorgos Dimitrakopoulos}
\IEEEauthorblockA{Integrated Circuits Lab, 
Electrical and Computer Engineering, 
Democritus University of Thrace, Xanthi, Greece
}}

\begin{document}

\maketitle

\begin{abstract}
Transformers and large language models (LLMs), powered by the attention mechanism, have transformed numerous AI applications, driving the need for specialized hardware accelerators. A major challenge in these accelerators is efficiently detecting errors caused by random hardware faults. Traditional algorithm-based fault tolerance (ABFT) techniques verify individual matrix multiplications but fall short in handling the full attention mechanism, particularly due to intermediate softmax normalization. This work proposes Flash-ABFT, a novel method that computes an online checksum across the entire three-matrix product of query, key and value matrices, of an attention layer, including the softmax operation, with a single check. This approach significantly reduces overhead by eliminating redundant checks while maintaining high fault-detection accuracy. Experimental results demonstrate that Flash-ABFT incurs only 5.3\% hardware area overhead and less than 1.9\% energy overhead, making it a cost-effective and robust solution for error detection in attention accelerators.
\end{abstract}


\section{Introduction}
Transformers are a deep learning model used for natural language processing~\cite{t5,llama} and computer vision tasks~\cite{vit}. A key innovation driving this progress is the attention mechanism~\cite{attention-is-all-you-need}, which allows models to focus on relevant parts of the input data. 

\begin{figure}[t]
\centering
\includegraphics[width=0.65\columnwidth]{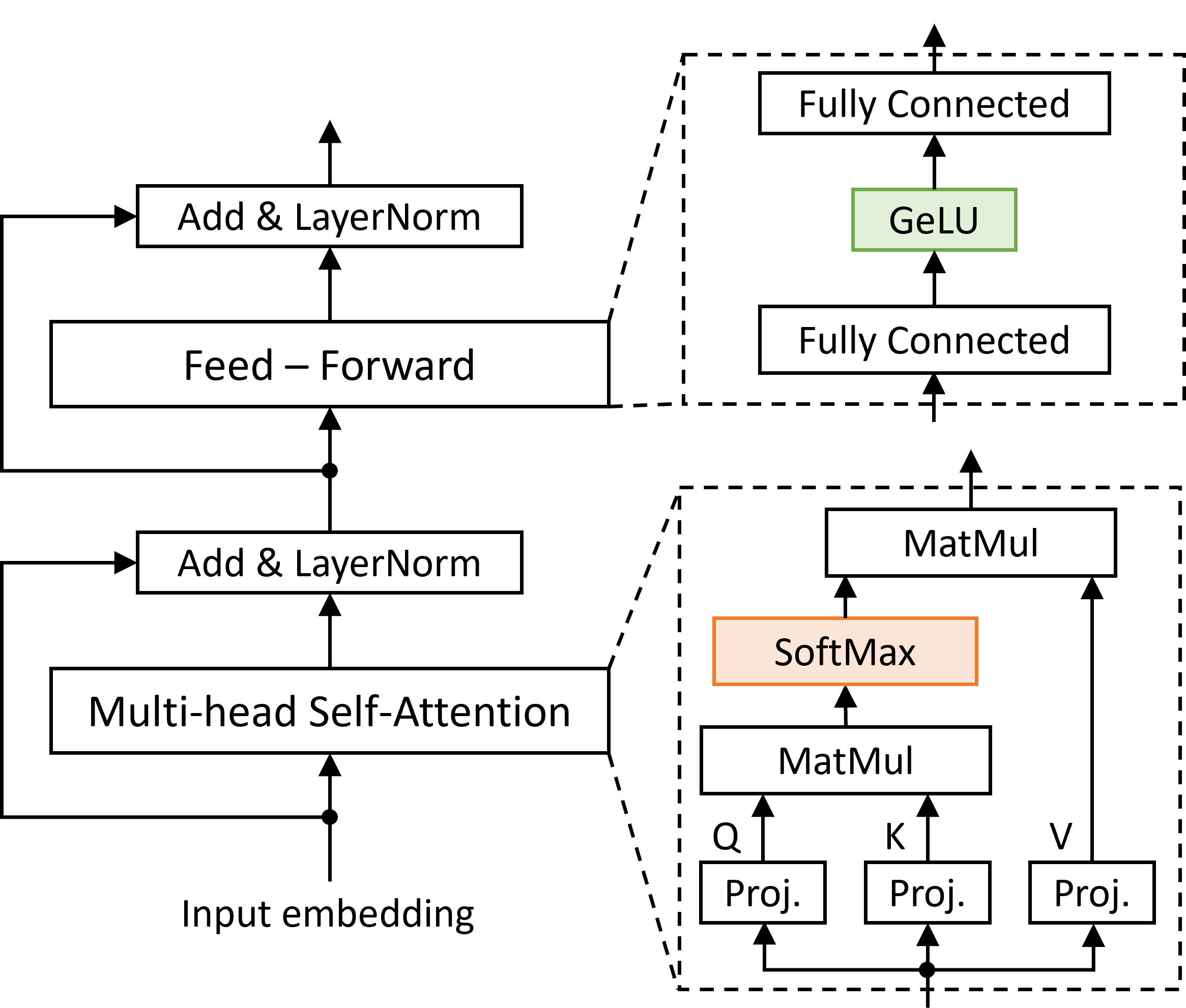}
\caption{An encoder-only transformer layer. For instance, BERT-base~\cite{bert} consists of twelve such layers.}
\label{f:tranformer}
\end{figure}

Transformer networks are composed of encoder and decoder blocks~\cite{attention-is-all-you-need} that include mainly matrix multiplications as well as softmax, normalization and GELU operations. Fig.~\ref{f:tranformer} depicts a layer of an encoder-only transformer ~\cite{bert,albert}. The input embedding is first projected to Query ($Q$), Key ($K$) and Value ($V$) matrices through a linear transformation. Then, $Q$ and $K$ matrices are multiplied and scaled to calculate, for each embedding, the importance of its neighbors. The result is passed through a softmax operator and the output is multiplied with matrix $V$ to compute the attention matrix. To complete self-attention the output is normalized and added to the input of the attention block. The self-attention block is followed by a feed-forward block that consists of two fully-connected layers that are separated by a GELU activation function. Decoder and encoder blocks follow a similar structure. Their difference is that the decoder consists of two self-attention blocks followed by a feed-forward block~\cite{attention-is-all-you-need}.

The growing need to process long sequences in transformer models has highlighted the complexity of the attention mechanism~\cite{longformer}. This computational challenge makes inference costly for extended context lengths. Traditional attention computes pairwise similarities between all tokens, resulting in a high number of operations and significant memory traffic. To alleviate this, techniques such as sparse attention~\cite{sparse_attn}, linear attention~\cite{lin_attn}, and low-rank attention~\cite{low_rank_attn_2020} have been explored. Each approach aims to reduce the number of operations by either approximating the full attention matrix or selectively focusing on the most relevant parts of the input sequence, striking a balance between accuracy and computational efficiency.

On the other hand, FlashAttention~\cite{fa,fa2,nsquared}, initially developed for GPUs, has become a highly effective method for accelerating attention. By utilizing tiling and combining online softmax computation with matrix operations, it facilitates parallel execution and minimizes memory traffic. These capabilities enable FlashAttention to reduce execution latency and streamline the processing of long sequences, all while maintaining accuracy.

In addition to performance and energy efficiency, modern accelerators must also ensure reliability~\cite{su2023testability}. This work tackles a crucial architectural challenge for attention accelerators: how to verify the online correctness of an attention kernel in the presence of random hardware faults, while minimizing the overhead of such checks.
Managing random hardware faults~\cite{soft-errors, unreliable} requires specialized hardware modules for fault detection~\cite{fault-tolerant-systems}. These faults should be detected online, ideally within a few cycles of their occurrence, to facilitate quick recovery. 

Algorithm-Based Fault Tolerance (ABFT)~\cite{abft} provides a cost-effective method for detecting errors in matrix-based computations~\cite{abft-pratical} by comparing the actual output checksum with a predicted one. ABFT has been also customized for specific computations such as CNNs~\cite{conv-checksum-tvlsi, making}, GCNs~\cite{gcn-abft} or attention layers of Transformers. In the latter case, attention computation is not checked as a whole. Instead, each matrix multiplication step involving the query, key, and value matrices is verified separately. In this context, ApproxABFT~\cite{approxabft} selectively detects and corrects significant errors during neural network inference. ATTNChecker~\cite{ATTNChecker} targets extreme errors for floating point such as INF, NaN, near-INF during training, while~\cite{ets} focuses on error vulnerability analysis by injecting faults into individual fully connected layers of Transformer networks.

In this work, we present Flash-ABFT, a custom-designed ABFT method tailored for attention layers. Unlike traditional approaches~\cite{ATTNChecker, approxabft}, which verify each matrix multiplication within the attention layer separately, Flash-ABFT computes a predicted checksum for the entire attention operation in a single, unified step. This fused checksum approach streamlines and simplifies error detection.

The contributions of this work can be summarized as follows:
\begin{itemize}
\item
We propose a method that fuses error checking into a single step by computing a predicted checksum for the entire attention operation, including the softmax normalization. Unlike traditional ABFT techniques that check each matrix multiplication separately, this unified approach enables more comprehensive and efficient error detection.

\item
This fused checksum computation can be seamlessly integrated into existing optimized hardware accelerators for attention, adding online error detection with minimal overhead. It enhances fault tolerance while incurring less than 1.9\% additional energy cost, making it a highly efficient and practical enhancement.

\item
Additional fault-injection experiments show that the proposed online checking logic provides high fault detection accuracy with minimal false alarms. This is achieved due to the small size of the check state relative to the accelerator's datapath.
\end{itemize}

\section{Attention Algorithm and FlashAttention}
The attention mechanism is a fundamental concept in machine learning, particularly in deep learning models that utilize the Transformer architecture~\cite{gpt2,vit}. 
The attention computation begins by multiplying the query 
$Q$ and key $K$ matrices, followed by scaling, to determine the relative importance of each embedding's neighbors. The resulting scores are then passed through a softmax function, and the output is multiplied by the value matrix $V$ to produce the final attention matrix. In summary, attention is computed as follows:
\begin{equation}
\text{attn}(Q, K, V)=\text{softmax}(Q\,K^T)\,V   
\label{e:attn}
\end{equation}

In practice, the attention mechanism operates across multiple heads in parallel, known as multi-head attention, allowing the model to capture complex relationships more effectively~\cite{base_attn}. This approach improves robustness against errors in generation and enables parallel execution across heads, enhancing overall performance. In the following, without loss of generality, we will limit our discussion to a single-head attention.

\begin{algorithm}
\caption{Attention with lazy softmax division}
\label{alg:attn-lazy}
\begin{algorithmic}[1]
\ForEach {query $\vec{q}$}
\For{$i = 1:N$} 
\State $s_i \gets \text{dot}(\vec{q}, \vec{k}_i)$
\State $m_i \gets \max(m_{i-1}, s_i)$
\EndFor
\State $\ell_0 \gets 0$
\For{$i = 1:N$} 
\State $\vec{o}_i \gets \vec{o}_{i-1} + e^{s_i - m_N}\cdot \vec{v}_i$
\State $\ell_i \gets \ell_{i-1} +e^{s_i - m_N}$
\EndFor
\State $\text{attn}(\vec{q}, K, V) \gets \vec{o}_N/\ell_N$
\EndFor
\end{algorithmic}
\end{algorithm}

Alg.~\ref{alg:attn-lazy} illustrates the computation of attention using the lazy softmax division concept~\cite{lazy_softmax, nsquared}. The process begins by computing the dot product between the query vector and all key vectors while identifying the maximum score for numerical stability. To prevent numerical instability caused by exponentiating large values, softmax is implemented by subtracting the maximum score from all values. This adjustment mitigates overflow issues while preserving the essential properties of the softmax function.
In the following, the output vector is computed incrementally as the weighted sum of each value value and the corresponding exponentiatial score. In parallel, the sum of all exponentials is also incrementally computed in $\ell_i$. The attention vector for one query vector $q$ is finalized by dividing the final output vector by the corresponding sum-of-exponents $\ell_N$~\cite{lazy_softmax, nsquared}.

The attention computation in Alg,~\ref{alg:attn-lazy} has a major bottleneck, as output calculation can only start after determining the maximum attention score, limiting efficiency for large sequence lengths 
($N$). FlashAttention~\cite{fa}, inspired by online softmax computation~\cite{online-softmax}, addresses this by merging the two inner loops of Alg,~\ref{alg:attn-lazy} into a single loop that computes all necessary variables online.

Alg,~\ref{alg:flash-attn2} presents FlashAttention-2~\cite{fa2}, an optimized variant of FlashAttention~\cite{fa}, that postpones the softmax division until the end using the sum of exponents~\cite{lazy_softmax, nsquared}. Unlike baseline attention, it computes all required variables within the same loop, eliminating the need to precompute the maximum attention score, which is crucial for maintaining efficiency with large sequence lengths.

\begin{algorithm}[t]
\caption{FlashAttention-2 with delayed softmax division}\label{alg:flash-attn2}
\begin{algorithmic}[1]
\ForEach {query $\vec{q}$}
\For{$i = 1:N$} 
\State $s_i \gets \text{dot}(\vec{q}, \vec{k}_i)$
\State $m_i \gets \max(m_{i-1}, s_i)$
\State $\ell_i \gets \ell_{i-1}e^{m_{i-1}-m_i}+e^{s_i-m_i}$
\State $\vec{o}_i \gets \vec{o}_{i-1} e^{m_{i-1}-m_i}+\vec{v}_i e^{s_i-m_i}$
\EndFor
\State $\text{attn}(\vec{q}, K, V) \gets \vec{o}_N/\ell_N$
\EndFor
\end{algorithmic}
\end{algorithm}

In each iteration, the dot product between the query vector and a key vector produces a similarity score $s_i$, and the current maximum attention score $m_i$ is updated. The accumulated sum of exponentials, $\ell_i$, is then computed by adjusting for the difference between the current and previous maximum values. The term $e^{m_{i-1}-m_i}$ ensures that $\ell_i$ remains properly updated, when $m_i$ surpasses the previous maximum $m_{i-1}$. Similarly, the output vector $\vec{o}_i$ is updated by weighting the new value vector $\vec{v}_i$ with its incremental softmax importance and adjusting the previous output $\vec{o}_{i-1}$. Finally, the attention output $\vec{o}_N$ is normalized by dividing it by the total accumulated sum of exponentials stored in $\ell_N$.

\begin{figure}
\centering
\includegraphics[width=0.98\columnwidth]{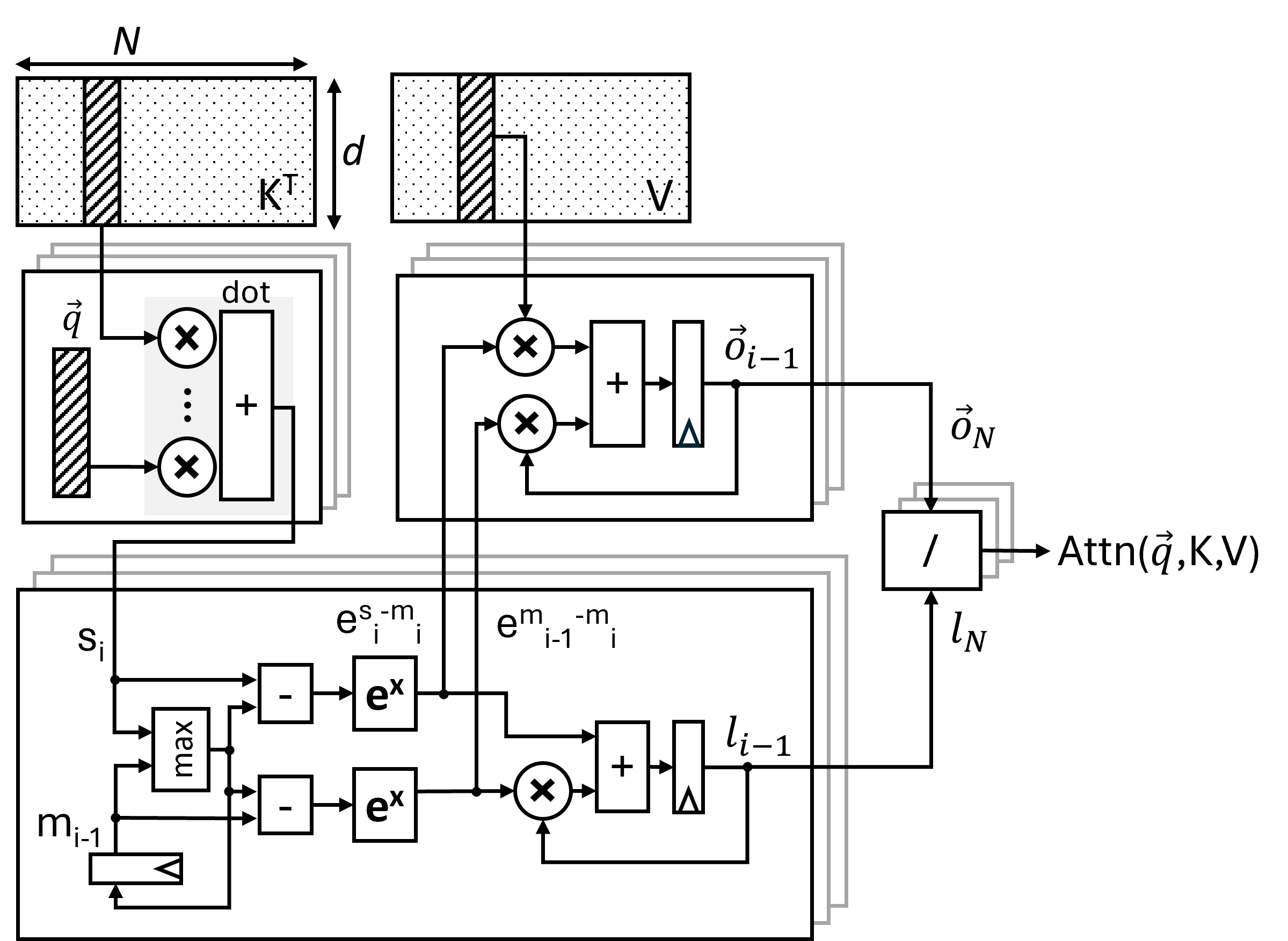}
\caption{A block-parallel hardware architecture for FlashAttention-2 kernel.}
\label{f:flashattn2-hw}
\end{figure}

The FlashAttention-2 kernel, as shown in Alg,~\ref{alg:flash-attn2}, utilizes two for loops that can be unrolled to improve parallelism and increase computational throughput. To eliminate serial dependencies, we unroll the outer loop, enabling the FlashAttention-2 kernel to process multiple query vectors concurrently within the same blocks of key and value vectors.
Figure~\ref{f:flashattn2-hw} illustrates the parallel hardware architecture derived for FlashAttention-2. In this setup, a block of query vectors is preloaded locally, while key vectors are sequentially fed to all blocks to compute the corresponding dot products. This process updates the maximum value and the running sum of exponentials. The output vectors for different query vectors are updated in parallel as value vectors are streamed into the accelerator. After all key and value vectors have been processed, the attention for each query vector is computed through a final division operation. The computation completes once all query vectors have been processed.

For FlashAttention-2 kernels, we assume that each cycle allows reading one key and one value vector, each containing 
$d$ elements, from local memory, where $d$ represents the size of the attention's hidden dimension.

\section{Attention-speficic ABFT}
ABFT offers a cost-efficient solution for detecting errors in matrix computations.
Adhering to the ABFT methodology~\cite{abft}, for validating the
matrix multiplication $C = A\, B$, it is essential to compare
the actual checksum of all elements of the output matrix $C$
with a predicted checksum derived independently by the elements of $A$ and $B$. The predicted checksum of $C$ can be computed by enhancing $A$ with an extra row that represents the per-column checksum vector of $A$ and matrix $B$ with an extra column that includes its per-row checksums and computing the dot product of the two checksum vectors.

For attention that is computed via~\eqref{e:attn}, we apply the same principle of ABFT, after appropriate computation reordering, by assuming that matrix $A$ corresponds to matrix $S = \text{softmax}(Q\, K^T)$ and $B$ to matrix $V$. First, we derive in Section~\ref{ss:form} the mathematical form of the attention-specific checksum prediction and then in Section~\ref{ss:online-check} we show how the computation of checksum can be computed online embedded with low-cost in FlashAttention hardware shown in Fig.~\ref{f:flashattn2-hw}.

\subsection{Attention Specific Checksum Prediction}
\label{ss:form}
To derive the needed attention-specific checksum prediction, we first need to examine the form of the output produced by a specific query. From~\eqref{e:attn}, the score of a query relative to a specific key vector is given by the dot product $s_{ij}=q_i \cdot k_j^T$, where $q_i$ represents the query vector, and $k_j^T$ corresponds to a key vector from the transposed matrix $K^T$.  

Performing the dot products across all columns of $K^T$,  we obtain a vector of scores $[s_{i1} \ s_{i2} \ ... \ s_{iN}]$ corresponding to the same query $q_i$, for a sequence length equal to $N$. Applying softmax to this vector of scores produces:

\begin{equation}
\begin{bmatrix}
e^{s_{i1}}/\sum_j e^{s_{ij}} & 
e^{s_{i2}}/\sum_j e^{s_{ij}} &
\ldots &
e^{s_{iN}}/\sum_j e^{s_{ij}}
\end{bmatrix}
\label{e:soft}
\end{equation}
In this softmax computation, for simplicity, we omitted the subtraction of the maximum score from all scores of the same row that is needed to preserve the numerical stability of softmax.

The vector shown in~\eqref{e:soft} corresponds to a row of the matrix $S$ that is going to be multiplied with matrix $V$ for computing attention. ABFT dictates
to compute a checksum for each column of $S$.
Summing all elements of the $k$th column of the normalized score matrix $\text{softmax}(Q\, K^T)$ leads to:
\begin{equation}
\label{e:sumcol}
sumcol_k(S)=\sum_{i=1}^N\frac{e^{s_{ik}}}{\sum_{j=1}^Ne^{s_{ij}}}, \text{for one $k$}
\end{equation}
Similarly for matrix $V$, we must compute a checksum vector that corresponds to sum of the elements of every row of $V$. The checksum for the $k$th row of $V$ is given by:
\begin{equation}
sumrow_k(V) = \sum_{j=1}^N v_{kj}
\label{e:sumrow}
\end{equation}

To compute the predicted checksum~\cite{abft}, we must compute the dot product between the per-column checksum vector of $S$, i.e., 
vector 
$[sumcol_1(S) \ sumcol_2(S) \ \ldots \ sumcol_N(S)]$ and the per-row checksum vector for matrix $V$, i.e., 
$[sumrow_1(V) \ sumrow_2(V) \ \ldots \ sumrow_N(V)]$

\begin{equation}
\label{e:check}
check=\sum_{k=1}^N sumcol_k(S) \cdot sumrow_k(V)    
\end{equation}
Replacing~\eqref{e:sumcol}  in~\eqref{e:check} we get 
\label{e:checkandcol}
\begin{equation}
check=\sum_{k=1}^N\left(\sum_{i=1}^N\frac{e^{s_{ik}}}{\sum_{j=1}^Ne^{s_{ij}}}\right) \cdot sumrow_k(V)       
\end{equation}
Since addition is both commutative and associative, and the number of summations is finite, the order of summation with respect to indices $i$ and $k$ can be interchanged, yielding that:
\begin{align}   
\label{e:checkandcol_inter}
check &= \sum_{i=1}^N\left(\sum_{k=1}^N\frac{e^{s_{ik}}}{\sum_{j=1}^N e^{s_{ij}}}\right) \cdot sumrow_k(V)\nonumber\\ 
&=\sum_{i=1}^N \left (\sum_{k=1}^N\frac{e^{s_{ik}}}{\sum_{j=1}^N e^{s_{ij}}} \cdot sumrow_k(V) \right)\nonumber\\ 
&=  \sum_{i=1}^N\frac{1}{\sum_{j=1}^N e^{s_{ij}}}
\left ( \sum_{k=1}^N e^{s_{ik}} \cdot sumrow_k(V) \right)
\end{align}

The change in the order of summation that resulted in \eqref{e:checkandcol_inter} allows us to compute the predicted checksum for the entire attention mechanism by summing 
$N$ independent checksums, each corresponding to a different query vector. This follows from \eqref{e:checkandcol_inter}, which allows us to express the relationship as
\begin{align}
check &= \sum_{i=1}^N check(q_i)\quad \text{with}\nonumber \\
check(q_i) &= \frac{1}{\sum_{j=1}^N e^{s_{ij}}}
\left ( \sum_{k=1}^N e^{s_{ik}} \cdot sumrow_k(V) \right)
\label{e:check-per-query}
\end{align}

\subsection{Integration of Checksum Computation in FlashAttention}
\label{ss:online-check}
As long as the predicted checksum can be derived independently per query using~\eqref{e:check-per-query}, it means that checksum computation can be embedded smoothly in the basic attention algorithm (Alg.~\ref{alg:attn-lazy}) or its FlashAttention-2 variant (Alg.~\ref{alg:flash-attn2}).

The term $\sum_{k} e^{s_{ik}}\cdot sumrow_k(V)$ that appears in the parenthesis in Eq.~\eqref{e:check-per-query} corresponds to the computation of the output in line 8 of Alg.~\ref{alg:attn-lazy}, where in place of each value vector $\vec{v}_k$, the sum of row elements $sumrow_k(V)$ is used. Also, the term $1/\sum_{j=1}^N e^{s_{ij}}$ corresponds to the sum of exponents of the same query vector computed in line 9 of Alg.~\ref{alg:attn-lazy} and used in the division of line 11 of the same algorithm. 
This analogy allows us to compute the predicted checksum recursively following an approach similar to Alg.~\ref{alg:attn-lazy} or Alg.~\ref{alg:flash-attn2}.

Enhancing the computation of FlashAttention-2 with online checksum computation that handles also adjustment of the incrementally computed maximum attention scores is shown in Alg.~\ref{alg:flash-attn2-abft} in blue. In line 7 of Alg,~\ref{alg:flash-attn2-abft}, the check per query $c_i$ is accumulated online ensuring that its old value, i.e., $c_{i-1}$, is properly adjusted when the maximum value is updated.
Once the computation for a query segment is complete, the check per query must be divided by the corresponding sum of exponents, as illustrated in line 10 of Alg,~\ref{alg:flash-attn2-abft}. In line 11 the checksums of different queries are accumulated.

\begin{algorithm}[h]
\caption{FlashAttention-2 with online checksum computation}\label{alg:flash-attn2-abft}
\begin{algorithmic}[1]
\ForEach {query $\vec{q}$}
\For{$i = 1:N$} 
\State $s_i \gets \text{dot}(\vec{q}, \vec{k}_i)$
\State $m_i \gets \max(m_{i-1}, s_i)$
\State $\ell_i \gets \ell_{i-1}e^{m_{i-1}-m_i}+e^{s_i-m_i}$
\State $\vec{o}_i \gets \vec{o}_{i-1} e^{m_{i-1}-m_i}+\vec{v}_i\cdot e^{s_i-m_i}$
\State {\color{blue} $c_i \gets c_{i-1}e^{m_{i-1}-m_i}+sumrow_i(V)\cdot e^{s_i-m_i}$}
\EndFor
\State $\text{attn}(\vec{q}, K, V) \gets \vec{o}_N/\ell_N$
\State {\color{blue} $check(\vec{q}) = c_N/\ell_N$}
\State {\color{blue} $check \gets check+check(\vec{q})$}
\EndFor
\end{algorithmic}
\end{algorithm}

The online computation of the predicted checksum shown in Alg.~\ref{alg:flash-attn2-abft} allows merging the incremental computation of per-query checksum (performed in line 7 of Alg.\ref{alg:flash-attn2-abft}) with the incremental computation of the output vector (line 6 in Alg.\ref{alg:flash-attn2-abft}). Both updates involve the same operations and can be written in a vector-merged form as follows:
\begin{equation}
\begin{bmatrix}
c_i \\           
\vec{o}_i
\end{bmatrix}=
\begin{bmatrix}
c_{i-1}\cdot e^{m_{i-1}-m_i}+sumrow_i(V)\cdot e^{s_i-m_i} \\           
\vec{o}_{i-1}\cdot e^{m_{i-1}-m_i}+\vec{v}_i\cdot e^{s_i-m_i}
\end{bmatrix}
\label{e:merged}
\end{equation}
Increasing by one element the output $\vec{o}_i$ and the value vector $\vec{v}_i$, i.e., $o^*_i = [c_i\quad \vec{o}_i]$ and $v^*_i=[sumrow_i(V)\quad \vec{v}_i]$ the merged incremental update of per-query checksum and the output shown in~\eqref{e:merged} can be written as:
\begin{equation}
o^*_i = o^*_{i-1}\cdot e^{m_{i-1}-m_i}+v^*_i\cdot e^{s_i-m_i}
\label{e:merged-output}
\end{equation}

\begin{figure}
\centering
\includegraphics[width=0.95\columnwidth]{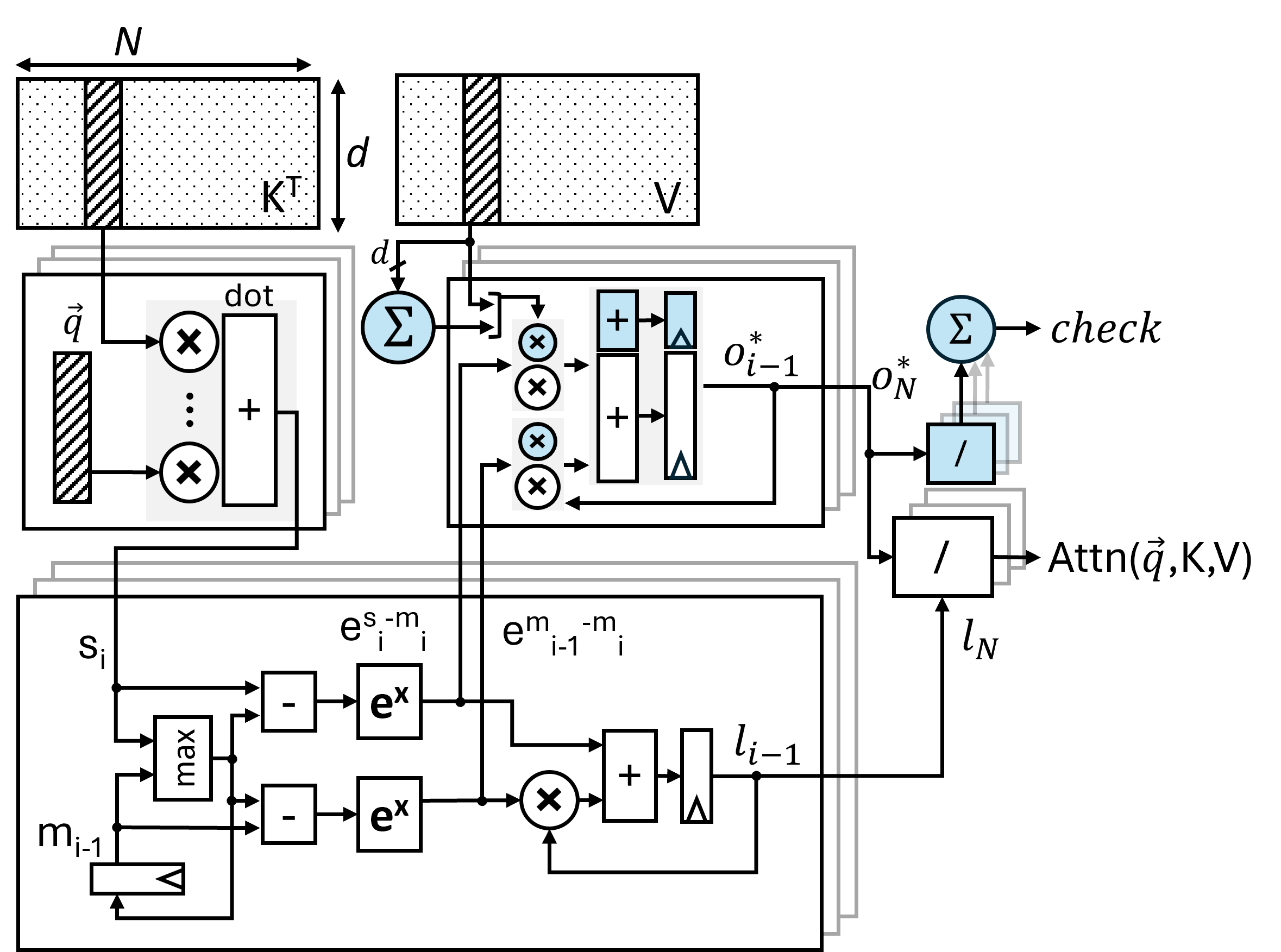}
\caption{A block-parallel hardware architecture for FlashAttention-2 kernel enhanced with online fault-detection logic that implements the proposed attention-specific ABFT.}
\label{f:flashattn2-abft-hw}
\end{figure}

This merging effectively allows to merge in hardware the computation of the output vector of FlashAttention-2 with the corresponding per-query checksum vectors. This is shown in Fig.~\ref{f:flashattn2-abft-hw} that highlights the way the per-query and across queries checksum is computed for a complete attention. An adder ($\sum$) computes the per-row checksum of matrix $V$, i.e., $sumrow_i(V)$, the enhanced output logic computes the per-query checks $c_i$, while the global dividers and accumulator complete the operations shown in lines 10 and 11 of Alg.~\ref{alg:flash-attn2-abft}.

\section{Evaluation}
The experimental results aim to highlight two key aspects of the proposed architecture. In the first set of experiments, we measure the hardware overhead caused by the error-checking logic in terms of area and power for a FlashAttention-based parallel hardware accelerator. In the second set of experiments, we examine the fault-detection capabilities of the custom attention checker.

\subsection{Hardware complexity evaluation}
The FlashAttention-based hardware accelerator including also the customized error-detection logic  has been described in C++ and synthesized to Verilog RTL using Catapult HLS. We examined two cases that correspond to the design depicted in Fig.~\ref{f:flashattn2-abft-hw} 
and serve 16 and 32 query vectors in parallel, respectively, assuming a hidden dimension of 128 elements as used in state-of-the-art transformers. The query vectors are preloaded separately in the accelerator, while the key and value vectors are loaded and broadcasted to all parallel blocks. Arithmetic operators inside the accelerator refer to reduced precision BFloat16 format, while all checksum accumulators are built with double-precision floats.

Clock frequency target was set to 500 MHz. The reported area results were derived from the Cadence synthesis tool driven by a 28-nm standard-cell library. Power estimation was done using PowerPro tool, while switching activity was derived by running attention kernels for various Large Language Models and benchmarks from PromptBench. Power estimation excludes memory power and focuses solely on the average power consumption of the computation kernel and the associated error checking logic. The memory power in not affected by the precense of the error-checking logic and is a result of the data flow imposed by FlashAttention.

\begin{figure}
    \centering
    \includegraphics[width=0.75\linewidth]{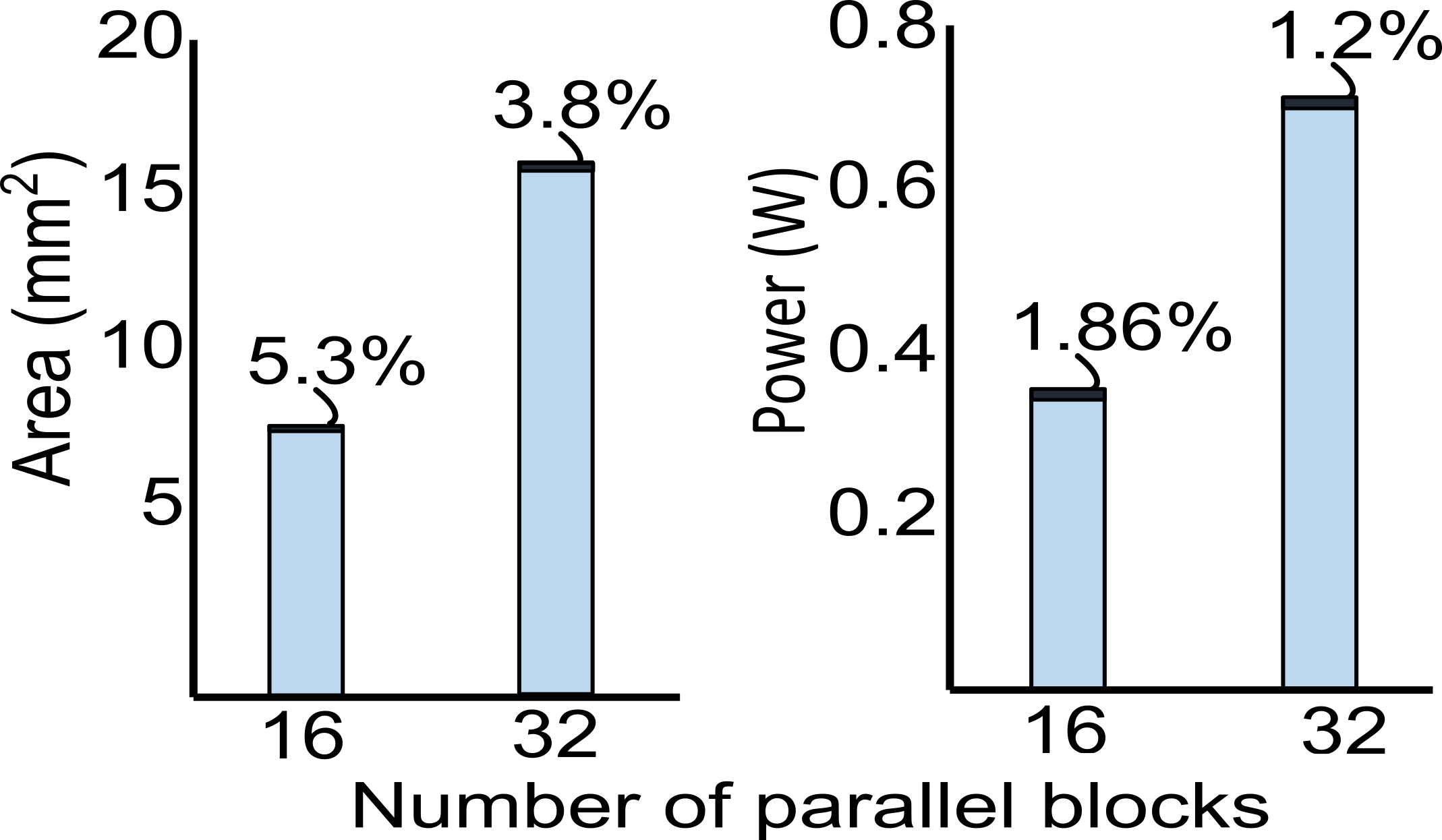}
    \caption{The area and average power consumption of the FlashAttention-2 accelerator extended with the proposed online error-detection logic at 28 nm, when computing attention for 16 and 32 query vectors in parallel, with hidden dimension $d = 128$. Both diagrams highlight separately the contribution of the checker’s logic.}
    \label{f:area_power}
\end{figure}

Fig~\ref{f:area_power} showcases that the additional area and power introduced by the proposed error detection logic for both designs is relatively small. The average area and power overhead is 4.55\% and 1.53\% respectively. Left checksum summation is shared across the blocks, as shown in Fig.~\ref{f:flashattn2-abft-hw}, thus making it contribute less to the total area overhead.

\subsection{Evaluation of fault detection properties}
To evaluate the error detection capability of the proposed checker, we inject a bit-flips in random clock cycles during the execution of an attention layer. 
Faults are injected to randomly selected storage elements covering both
the registers of the FlashAttention-2 kernel and the registers of the checking logic. 
Within a register each bit has an equal probability of being flipped.
Whether a fault will be injected on the FlashAttention-2 hardware or the checker depends on the amount of their storage elements. A fault is more probable to hit the
FlashAttention-2 hardware than the checker’s logic since it includes
significantly more storage elements including the locally stored query and output vectors as well as registers for storing the maximum attention score and running sum-of-exponents for each query.
Also, we assume that memory that stores query, key and value matrices before being loaded to the accelerator is protected by a separate error detection logic and thus input data fetched are fault-free.

We injected faults to the first attention layer of four LLMs with different hidden dimensions using the same embedding prompt with sequence length of 256. 
Specifically, we evaluated the layers of Bert, Phi-3-mini, Llama-3.1, and Gemma2, which have hidden dimensions of 64, 96, 128, and 256, respectively and are available on HuggingFace~\cite{hf}. In this way, we can examine the impact of the hidden dimension on the detection rate.
The observed behavior fell into one of the following three categories:

\begin{enumerate}
\item Detected: A faulty output was generated, and the proposed ABFT-based checking logic successfully identified it.

\item False Positive: A fault was injected into one checksum accumulator either the local ones that compute checksums per query vector or the global one that adds the checksum of all parallel queries.
This led ABFT to incorrectly classify a correct output as erroneous.

\item Silent: The error remains undetected by ABFT either due to rounding or because the bit flip resulted in invalid floating point numbers such as NaN~\cite{float_nan}.
\end{enumerate}

To prevent silent faults due to rounding during our fault-injection campaigns, we consider a fault detected if the predicted checksum differs by the true output checksum by more than $10^{-6}$. We found this limit out experimentally for the examined attention layers.
False negative faults require a fault injected to matrix
multiplication and checksum accumulation to cancel each
other thus causing ABFT to fail to identify an actual fault in
the output. We couldn't identify such cases in our experiments.

\begin{table}[t!]
\centering
\caption{Fault detection accuracy for a single injected fault using an error bound of $10^{-6}$ .}
\label{t:detection_rate}
\begin{tabular}{|c||c|c|c|c|}
\hline
\multirow{2}{*}{Fault injection behavior}&\multicolumn{4}{c|}{Sequence length=256}\\\cline{2-5}

&d=64&d=96&d=128&d=256\\
\hline
Detected&96.94\%&97.56\%&98.45\%&98.87\%\\
False Positive&2.66\%&1.99\%&1.25\%&0.62\%\\
Silent&0.4\%&0.45\%&0.3\%&0.51\%\\
\hline
\end{tabular}
\end{table}

The results of injecting a single fault, based on 10,000 independent fault-injection campaigns, are summarized in Table~\ref{t:detection_rate}. Conducting additional fault-injection campaigns does not change the observed behavior. In this setup, a single fault was injected into each application, assuming an input sequence length of 256 elements.

As shown, FlashAttention-ABFT successfully detects faults in most cases. However, silent faults still occur due to bit errors that alter numerical values in a way that makes them unrepresentable within the rules of floating-point arithmetic.

The occurrence of false checker reactions is minimal, reflecting the low probability that a fault will affect the checker rather than the FlashAttention-2 kernel. Furthermore, it is clear that as the hidden dimension increases, the detection rate improves. A larger hidden dimension increases the number of elements stored in the query and output vector registers, thereby reducing the likelihood of a fault affecting the checker's logic and causing false positive fault injections.

As the number of injected faults per fault-injection campaign increases (1–5 faults are randomly injected) the observed results change significantly and the possibility of having a false alarm is almost zero on average.

Finally, it should be noted that the fault detection percentages reported correspond to the arithmetic discrepancies identified by the proposed ABFT logic at the checksum level. If the injected faults are actually critical for the overall performance of the LLM application is not quantified and is part of future work. 

\section{Conclusions}
The proposed ABFT technique for attention achieves a cost-effective approach to error detection by calculating an online checksum for the entire attention process, including the softmax normalization. This advancement is crucial because it allows for the detection of errors caused by random hardware faults in attention accelerators. The technique is particularly important as it addresses a gap in existing algorithm-based fault tolerance (ABFT) methods, which can only verify individual matrix multiplications but fail to check the entire attention mechanism. Furthermore, experimental results demonstrate that the method incurs minimal hardware area and energy overhead, while maintaining high fault-detection accuracy, as shown by the presented fault-injection analysis. This makes the approach both efficient and reliable, enhancing the performance of specialized hardware accelerators such as those using FlashAttention variants.

\bibliographystyle{IEEEtran}
\bibliography{refs}

\end{document}